\documentclass[journal]{IEEEtran}

\IEEEoverridecommandlockouts                             
\usepackage{tikz}
\usepackage{graphicx, subfigure} 
\usepackage{epsfig} 
\usepackage{mathptmx} 
\usepackage{times} 
\usepackage{amsmath} 
\usepackage{amssymb}  
\usepackage{url}
\usepackage[lined,boxed,commentsnumbered, ruled]{algorithm2e}
\usepackage{cite}
\usepackage{xcolor,colortbl}
\definecolor{Lightcyan}{rgb}{0.75,1,1}
\definecolor{Lightyellow}{rgb}{1,1,0.85}
\definecolor{Red}{rgb}{0.75,0,0}
\definecolor{Green}{rgb}{0,0.75,0}
\usepackage{soul}
\usepackage[firstpage]{draftwatermark}
\SetWatermarkText{This paper has been accepted for publication at IEEE 24th Intelligent Transportation Systems Conference (ITSC) Indianapolis United States 2021 Copyright IEEE}
\SetWatermarkVerCenter{1.5cm}
\SetWatermarkAngle{0}
\SetWatermarkFontSize{8pt}
\SetWatermarkColor[gray]{0.4}
\SetWatermarkScale{0.9}

\title{\LARGE \bf Piecewise Linear De-skewing for LiDAR Inertial Odometry}

\author{John Henawy$^{1,*}$, Zhengguo Li$^{1,*}$, Wei Yun Yau$^{1}$, Gerald Seet$^{2}$, and Kong Wah Wan$^{1}$
\thanks{$^*$Joint first author.}
\thanks{$^1$John Henawy, Zhengguo Li, Kong Wah Wan, and Wei Yun Yau are with the Institute for Infocomm Research, A-STAR, Singapore, 138632
        {\tt\small \{john\_henawy, ezgli, kongwah, wyyau\}@i2r.a-star.edu.sg}}
\thanks{$^2$ Gerald Seet is with the School of Mechanical and Aerospace
Engineering, Nanyang Technological University, Singapore, 639798, MGLSEET@ntu.edu.sg.}
}

\begin{document}

\maketitle
\thispagestyle{empty}
\pagestyle{empty}

\begin{abstract}
Light detection and ranging (LiDAR) on a moving agent could suffer from motion distortion due to simultaneous rotation of the LiDAR and fast movement of  the agent.  An accurate piecewise linear de-skewing algorithm is proposed to correct the motion distortions for LiDAR inertial odometry (LIO) using  high frequency motion information provided by an Inertial Measurement Unit (IMU).  Experimental results show that the proposed algorithm can be adopted to improve the performance of existing LIO algorithms especially in cases of fast movement.
\end{abstract}


\section{Introduction}
On-line localization of mobile agents plays an important role in autonomous navigation and it has been intensively studied for many years \cite{henawy2019accurate,demir2019robust,cao2020accurate,javanmardi2017autonomous,chen2019aerial}. Since a light detection and ranging (LiDAR) rangefinder can provide highly reliable and precise distance measurements for surrounding environments and it is insensitive to ambient lighting and optical texture in a scene, the LiDAR rangefinder is one of the most widely adopted sensors for the indoor navigation among all available sensors \cite{shan2018lego,le2019in2lama,chow2019toward}.  Pose of a rangefinder can be estimated by matching two different scans \cite{Lu1997robot}. Iterative Closest Point (ICP) algorithm is one of the dominant solutions for the scan matching problem \cite{chen1992object,besl1992method}. In existing ICP algorithms \cite{chen1992object,besl1992method,rusinkiewicz2001efficient,minguez2006metric,diosi2007fast},  the rotation of a laser beam  is assumed to be the only motion of a scanning device. However, the rangefinder mounted on an autonomous vehicle or an unmanned aerial vehicle (UAV) usually moves during scanning a sweep due to movement of the agent. This introduces motion distortions among all points in the sweep, especially when the agent is moving fast.

Distortion caused by the movement of the agent affects accuracy of LiDAR odometry (LO). This problem was addressed by Bezet and Cherfaoui \cite{bezet2006time} in terms of time error correction. It is assumed that the sensor measures the distance to the same object between two frame at a same angle.  The algorithm in  \cite{bezet2006time} is not applicable if the agent rotates and moves fast. Zhang \textit{et al}. \cite{zhang2014loam} proposed an interesting LO algorithm by computing the transformation matrix from the first point to the last point of a sweep and linearly interpolating a transformation matrix to correct motion distortion for each point in the sweep.   Accurate LO requires knowledge of the LiDAR pose during continuous laser ranging to correct the motion distortion. Fortunately, the knowledge can be provided by an IMU. The linear acceleration from IMU was also utilized in \cite{zhang2014loam} to address the nonlinear motion of the agent. Experimental results indicate that the performance of the ICP can indeed be improved using the IMU measurements. Ye \textit{et al}. \cite{ye2019tightly} proposed an interesting de-skewing algorithm by using the IMU measurements. The transformation matrix from the first point to the last point of a sweep was first computed by using all the IMU measurements during the sweep. A transformation matrix is then linearly interpolated to correct the motion distortion for each point in the sweep. Same as \cite{zhang2014loam}, the movement of the agent was assumed to be linear during the sweep. The linear assumption  is not true if the motion of the agent is complicated. For example, a sweep can be scanned when the agent goes straight and followed by a turn. It is desired to develop a new de-skewing algorithm without the linear assumption required by \cite{zhang2014loam,ye2019tightly}. Moreover, high accuracy preintegrated IMU models were proposed recently \cite{henawy2019accurate,henawy2020accurate,eckenhoff2019closed,forster2016manifold}. All these models are derived for the moving agents where IMU pre-integration is considered in the world reference frame. On the other hand, the skewing  distortion of 3D points occurs on the body frame. Therefore, the IMU pre-integration of 3D points in the body frame is worthy of being studied when the de-skewing of LiDAR scans is studied.

In this paper, an accurate piecewise linear de-skewing algorithm is proposed to correct motion distortions caused by the movement of the agent. The proposed algorithm is based on an observation that the IMU has a potential to provide the knowledge of the LiDAR pose during continuous laser ranging due to its high sampling rate. An elegant discrete model is first derived for a static 3D point of the environment in the body frame under the same assumption in \cite{henawy2019accurate,henawy2020accurate,eckenhoff2019closed}, i.e., the linear acceleration and the angular velocity in the IMU frame are constant between two successive IMU measurements. The derivations in the body frame are much simpler than those in the world reference frame \cite{henawy2019accurate,henawy2020accurate,eckenhoff2019closed}. The new model is applied to derive a closed form IMU integration model for the static 3D point in the body frame. The proposed IMU motion integration model is different from the models in \cite{henawy2019accurate,henawy2020accurate,eckenhoff2019closed} in the sense that only the proposed model is on the static 3D point. Finally the proposed IMU integration model is adopted to correct the motion distortions caused by the movement of the agent. One point in a sweep is selected as the referencing point and each point in the sweep is corrected by using a transformation matrix which is computed by using all the IMU measurements between the referencing point and the point. Clearly, the proposed algorithm does not assume the motion of the agent to be linear during the sweep as required by \cite{zhang2014loam,ye2019tightly}. The corrected scans serve as the inputs of an existing LIO such as \cite{ye2019tightly}.  The proposed motion integration model is integrated with  the LiDAR factor via a joint optimization  with the mean measurement, the covariance matrix and the Jacobian matrix of the proposed model in  \cite{henawy2020accurate}. As such, the proposed piecewise linear de-skewing algorithm and the LIO are seamlessly integrated together to reduce the motion distortion caused by the movement of the agent. The proposed de-skewing algorithm is  complimentary to the existing LIO algorithms such as \cite{ye2019tightly} and it can be used to improved the robustness of the existing LIO  algorithms.  Overall, two major contributions of this paper are: 1) an elegant closed form IMU integration model in the body frame for the static 3D point by using the IMU measurements, and 2) a piecewise linear de-skewing algorithm for correcting the motion distortion of the LiDAR which can be adopted by any existing LIO algorithm.

The rest of the paper is organized as follows. A literature review of  related works is provided in Section \ref{Related}. Section \ref{initialvaluesection} includes an IMU integration model for the 3D point, and section \ref{proposed_deskew} introduces the proposed de-skewing algorithm. Extensive results are given in Section \ref{experimentalresults} to verify the proposed algorithm. Finally, some conclusion remarks are provided in Section \ref{conclusion}.

\section{Related Works}
\label{Related}
\begin{figure}[!htb]
	\centering{
\includegraphics[width=2.6in]{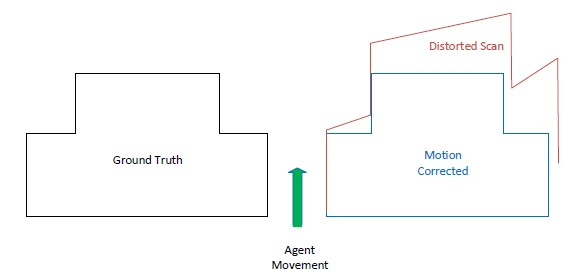}}
\caption{Motion distortions in a sweep of LiDAR. Black points is a given an environment, red points stand for a distorted scan, and blue points are a motion corrected scan.}
	\label{LiDAR_Distortion}
\end{figure}

The existing ICP algorithms \cite{chen1992object,besl1992method,rusinkiewicz2001efficient,minguez2006metric,diosi2007fast,pomerleau2013comparing} assume the rotation of a laser beam  to be the only motion of a scanning device. Unfortunately, this is not always true \cite{hong2010vicp}. An example is illustrated in Figure \ref{LiDAR_Distortion}. The black line in Figure \ref{LiDAR_Distortion} represents an environment and the movement of the rangefinder is indicated by the green arrow while the red points are the raw scan data. Obviously, the set of the points is distorted due to the movement of the agent.

Accurate LO requires knowledge of the LiDAR pose during continuous laser ranging to correct the motion distortion, and the knowledge can be provided by an IMU. The LiDAR and IMU measurements could be loosely coupled. Zhang \textit{et al}. \cite{zhang2014loam,zhang2017low} calculated the orientation of the robot pose by IMU measurements in LiDAR odometry and mapping (LOAM) algorithm. The LOAM considers the IMU measurements as a prior factor only for the algorithm. As a result, the IMU measurements were not used in the optimization process. Tang \textit{et al}. \cite{tang2015lidar} proposed to fuse the LiDAR and IMU measurements using an Extended Kalman filter (EKF). However, this algorithm was proposed for 2D scenarios and it is not applicable for a 3D environment. Lynen \textit{et al}. \cite{lynen2013robust} proposed a robust multi-sensors fusion approach  which is applicable to the 3D environment. The LiDAR and IMU measurements were also fused using the EKF. Generally, the loosely coupled paradigms are simpler but less accurate.

The LiDAR  and IMU measurements can also be tightly coupled. Soloviev \textit{et al}. \cite{soloviev2007tight} proposed an algorithm for 2D LiDAR scanning and matching. IMU orientation measurements were used to compensate the tilted LiDAR. Moreover, the drift in IMU measurements was corrected and updated using the Kalman filter (KF). One more tightly coupled algorithm was proposed by Hemann \textit{et al}. \cite{hemann2016long}. The algorithm can be applied to achieve good and robust results over long travelling period. However, it could fail without prior information about the environment.  Inspired by the work in \cite{qin2018vins}, Ye \textit{et al}. \cite{ye2019tightly} proposed a 3D tightly coupled fusion algorithm between the LiDAR and IMU measurements. One common way to address the motion distortion is the continuous-time representation such as the work in \cite{park2018elastic,le20183d}. Another technique is the linear interpolation approach \cite{zhang2014loam}. The movement of the agent was assumed to be linear during the sweep. Its accuracy could be hindered by the linear motion assumption. It is thus desired to design a new de-skewing algorithm  in order to compensate the motion distortion by using the high frequency IMU measurements.

\section{IMU Pre-integration of 3D Points}
\label{initialvaluesection}
In this section, a new discrete motion pre-integration model is introduced for static 3D points of the environment in the LiDAR frame. Same as \cite{henawy2019accurate,henawy2020accurate}, the proposed model assumes that  both the linear acceleration and the angular velocity in the IMU frame are constant between two successive IMU measurements. It is worth noting that $a_b$ and $\omega_b$ can be derived by multiplying the readouts of the IMU by the fixed rotational matrix between the IMU and LiDAR. All notations of the proposed model are included in Table \ref{table2}.

{\small \begin{table}[!t]
	\centering
	\caption{Notations used in the proposed motion pre-integration model}
	\begin{tabular}{|c|c|}
		\hline
	 $V_b=\left[\begin{array}{lll} V_x & V_y & V_z\end{array}\right]^T$     & linear velocity of the \\ & agent
in the LIDAR frame \\ \hline
$R_{bw}=\left[\begin{array}{lll} R_1 & R_2 & R_3\end{array}\right]$  & rotation matrix from the world \\
& frame to the LIDAR frame \\ \hline
$a_b$ &  linear acceleration of the agent \\
&  in the LIDAR frame\\ \hline
$g$ &  gravitational acceleration \\		\hline
$\omega_b=\left[\begin{array}{lll} \omega_x & \omega_y & \omega_z\end{array}\right]^T$ & angular
velocity of the agent \\
& in the LIDAR frame\\ \hline
$P_{b,l} = \left[\begin{array}{lll} X_{l} &
 Y_{l} & Z_{l}\end{array}\right]^T$ &   the $l$th 3D\\
 & point in the LIDAR frame\\ \hline
 $\left[\begin{array}{lll} \hat{X}_{l} & \hat{Y}_{l} & \hat{Z}_{l}\end{array}\right]^T$ &  the $l$th synchronized 3D \\
 & point  in the LIDAR frame\\ \hline
   $\varsigma $ & duration between two\\
  & successive IMU measurements\\\hline
 		\end{tabular}
    \label{table2}
\end{table}}

The dynamics of the $l$th static 3D point of the environment is  represented as

{\small \begin{equation}
 \label{agentdynamics}
\left[\begin{array}{l}
\dot{P}_{b,l}\\
\dot{V}_b\\
\dot{R}_1\\
\dot{R}_2\\
\dot{R}_3\\
\end{array}
\right]=A(\omega_b)\left[\begin{array}{l}
P_{b,l}\\
V_b\\
R_1\\
R_2\\
R_3\\
\end{array}
\right]+\left[\begin{array}{l}
0\\
a_b\\
0\\
0\\
0\\
\end{array}
\right],
\end{equation}}

where  the matrix $A(\omega_b)$ is given by

{\small \begin{equation}
\label{Amatrix}
A(\omega_b) = \left[\begin{array}{lllll}-\omega_b^{\wedge} & -I & 0 & 0 & 0\\
0 & -\omega_b^{\wedge} & 0 & 0 & gI\\
0 & 0 & -\omega_b^{\wedge} & 0 & 0\\
0 & 0 & 0 & -\omega_b^{\wedge} & 0 \\
0 & 0 & 0 & 0 & -\omega_b^{\wedge}\\
\end{array}
\right],
\end{equation}}

$I$ is a $3\times 3$ identity matrix,  and $\omega_b^{\wedge}$ is defined as

{\small \begin{equation}
\omega_b^{\wedge}=\left[\begin{array}{lll}
0 & -\omega_z & \omega_y\\
\omega_z & 0 & -\omega_x\\
-\omega_y & \omega_x & 0\\
\end{array}
\right].
\end{equation}}

Since both $\omega_b$ and $a_b$ are constant in the interval between the two successive IMU measurements,  the system (\ref{agentdynamics}) is a time invariant linear system in the interval. It can be derived from the equation (\ref{agentdynamics}) that

{\scriptsize\begin{align}
\label{agentdynamicsrr}
\left[\begin{array}{l}
 P_{b,l}(k+1)\\
 V_b(k+1)\\
 R_1(k+1)\\
 R_2(k+1)\\
 R_3(k+1)\\
 \end{array}
 \right]=\exp^{A(\omega_b)\varsigma}\left[\begin{array}{l}
 P_{b,l}(k)\\
 V_b(k)\\
 R_1(k)\\
 R_2(k)\\
 R_3(k)\\
 \end{array}
 \right]+\int_{0}^{\varsigma}\exp^{A(\omega_b)\tau}d\tau\left[\begin{array}{l}
0\\
a_b\\
0\\
0\\
0\\
\end{array}
\right].
\end{align}}

Notice that the matrix $A(\omega_b)$ can be decomposed as

{\small \begin{equation}
A(\omega_b)=A_1(\omega_b)+A_2,
\end{equation}}

where the matrices $A_1(\omega_b)$ and $A_2$ are

{\small \begin{align}\left\{\begin{array}{l}
A_1(\omega_b)  = -\left[\begin{array}{lllll}\omega_b^{\wedge} & 0 & 0 & 0 & 0\\
0 & \omega_b^{\wedge} & 0 & 0 & 0\\
0 & 0 & \omega_b^{\wedge} & 0 & 0\\
0 & 0 & 0 & \omega_b^{\wedge} & 0 \\
0 & 0 & 0 & 0 & \omega_b^{\wedge}\\
\end{array}
\right]\\
A_2  = \left[\begin{array}{lllll}0& -I & 0 & 0 & 0\\
0 & 0 & 0 & 0 & gI\\
0 & 0 & 0 & 0 & 0\\
0 & 0 & 0 & 0 & 0 \\
0 & 0 & 0 & 0 & 0\\
\end{array}
\right]
\end{array}
\right..
\end{align}}

Since $A_1(\omega_b)A_2=A_2A_1(\omega_b)$, it follows that \cite{zhang2011matrix}

{\small \begin{align}
\exp^{A(\omega_b)\varsigma}=\exp^{A_1(\omega_b)\varsigma}\exp^{A_2\varsigma}.
\end{align}}

Two exponential matrices $\exp^{A_1(\omega_b)\varsigma}$ and $\exp^{A_2\varsigma}$ can be easily computed as \cite{zhang2011matrix}

{\small \begin{align}\left\{\begin{array}{l}
\exp^{A_1\varsigma}  = \left[\begin{array}{lllll}E(-\theta_b) & 0 & 0 & 0 & 0\\
0 & E(-\theta_b) & 0 & 0 & 0\\
0 & 0 & E(-\theta_b) & 0 & 0\\
0 & 0 & 0 & E(-\theta_b) & 0 \\
0 & 0 & 0 & 0 & E(-\theta_b)\\
\end{array}
\right]\\
\exp^{A_2\varsigma}  = \left[\begin{array}{lllll}I& -\varsigma I & 0 & 0 & -\frac{g\varsigma^2}{2}I\\
0 & I & 0 & 0 & g\varsigma I\\
0 & 0 & I & 0 & 0\\
0 & 0 & 0 & I & 0 \\
0 & 0 & 0 & 0 & I\\
\end{array}
\right]
\end{array}
\right.
\end{align}}

where $\theta_b$ is $\omega_b\varsigma$. The matrix $E(\theta_b)$ is defined as 

\begin{align}
E(\theta_b)=Exp(\theta_b)=I+\frac{s}{\|\theta_b\|}\theta_b^{\wedge}+\frac{1-c}{\|\theta_b\|^2}(\theta_b^{\wedge})^2
\end{align}

and $c=\cos\|\theta_b\|$ and $s=\sin\|\theta_b\|$.

Subsequently, $\exp^{A(\omega_b)\varsigma}$ can be computed as 

{\small \begin{equation}
\label{AT}
\left[\begin{array}{lllll}E(-\theta_b) & -\varsigma  E(-\theta_b) & 0 & 0 & -\frac{g\varsigma^2E(-\theta_b)}{2}\\
0 & E(-\theta_b) & 0 & 0 & g\varsigma  E(-\theta_b)\\
0 & 0 & E(-\theta_b) & 0 & 0\\
0 & 0 & 0 & E(-\theta_b) & 0 \\
0 & 0 & 0 & 0 & E(-\theta_b)\\
\end{array}
\right].
\end{equation}}

Due to the new states in the equation (\ref{agentdynamics}), the derivation of $\exp^{A(\omega_b)}$ is much simpler than the corresponding derivations in \cite{henawy2019accurate,henawy2020accurate,eckenhoff2019closed}. A discrete motion model between the two successive IMU measurements can be derived for the $l$th point as

{\small \begin{align}
\label{Vkequation}
\left\{\begin{array}{l}
P_{b,l}(k+1)=E(-\theta_b(k))P_{b,l}(k)-\varsigma E(-\theta_b(k)) V_b(k)-\\
\hspace{19mm} \frac{g}{2}\varsigma ^2 E(-\theta_b(k))R_3(k)-\Upsilon(-\theta_b(k))\hat{P}_b(k)\\
V_b(k+1) = E(-\theta_b(k))V_b(k)+g\varsigma E(-\theta_b(k)) R_3(k)\\
\hspace{17mm}+J_r(-\theta_b(k))\hat{V}_b(k)\\
R_{bw}(k+1)=E(-\theta_b(k))R_{bw}(k)
\end{array}
\right.,
\end{align}}

where $\hat{V}_b$, and $\hat{P}_b$ are $a_b\varsigma$, and $a_b\varsigma^2$, respectively. The matrices $J_r(\theta_b)$, and $\Upsilon(\theta_b)$ are defined as

{\small  \begin{align}
 \left\{\begin{array}{l}
 J_r(\theta_b)=I+\frac{1-c}{\|\theta_b\|^2}\theta_b^{\wedge}+\frac{\|\theta_b\|-s}{\|\theta_b\|^3}(\theta_b^{\wedge})^2\\
 \Upsilon(\theta_b)=\frac{I}{2}+\frac{s-\|\theta_b\|c}{\|\theta_b\|^3}\theta_b^{\wedge}+\frac{1-c+\frac{\|\theta_b\|^2}{2}-\|\theta_b\|s}{\|\theta_b\|^4}(\theta_b^{\wedge})^2
 \end{array}
 \right..
 \end{align}}

\section{The Proposed Piecewise Linear De-skewing}
\label{proposed_deskew}

Consider two time instances $k=i$ and $k=j$. It should be pointed out that  $\omega_b$ and $a_b$ in the interval between the $(k+1)$th and $(k+2)$th IMU measurements are different from those in the interval between the $k$th and $(k+1)$th IMU measurements. Therefore, the dynamics of the 3D point is a switched system   between the two time instances \cite{li2001robust,li2005switched}.

By accumulating $k$ from $i$ to $(j-1)$ and using the following two equations

\begin{align}
 \left\{\begin{array}{l}
 E(\theta_b(k))J_r(-\theta_b(k))=J_r(\theta_b(k))\\
E(\theta_b(k))\Upsilon(-\theta_b(k))=\Lambda(\theta_b(k))
 \end{array}
 \right.,
\end{align}

the discrete motion pre-integration model between the two time instances can  be derived for the $l$th 3D point as

{\small \begin{align}
\label{motionintegration}
\left\{\begin{array}{l}
P_{b,l}(j) = F^T(i,j)P_{b,l}(i)+t(i,j)\\
V_b(j) =F^T(i,j)(V_b(i)+g{\displaystyle \sum_{k=i}^{j-1}}\varsigma R_3(i)+\mu_b(i,j))\\
R_{bw}(j)=F^T(i,j)R_{bw}(i)
\end{array}
\right.,
\end{align}}

where $t(i,j)$ is a translator vector, $F(i,j)$ is a rotation matrix. $t(i,j)$, $F(i,j)$ and $\mu_b(i,j)$ are given as

{\small \begin{align}
\label{motionintegration2}
\left\{\begin{array}{l}
t(i,j)=-F^T(i,j)(\chi_b(i,j)+\zeta_b(i,j))\\
F(i,j)={\displaystyle \prod_{k=i}^{j-1}}E(\theta_b(k))\\
\chi_b(i,j)={\displaystyle \sum_{k=i}^{j-1}}\varsigma V_b(i)+\frac{g}{2}({\displaystyle \sum_{k=i}^{j-1}}\varsigma)^2R_3(i)\\
\zeta_b(i,j)={\displaystyle\sum_{k=i}^{j-1}}(F(i,k)\Lambda(\theta_b(k))\hat{P}_b(k)+\mu_b(i, k)\varsigma)\\
\mu_b(i,j)={\displaystyle\sum_{k=i}^{j-1}}F(i,k)J_r(\theta_b(k))\hat{V}_b(k)
\end{array}
\right.,
\end{align}}

$F(k,k)=I$ and $\mu_b(k,k)=0$ for all $k$'s, and the matrix $\Lambda(\theta_b)$ is

{\small \begin{align}
\Lambda(\theta_b)=\frac{1}{2}I+\frac{\theta_b-s}{\|\theta_b\|^3}\theta^{\wedge}+\frac{2c-2+\|\theta_b\|^2}{\|\theta_b\|^4}(\theta_b^{\wedge})^2.
\end{align}}

It is worth noting that the proposed IMU motion integration model is different from the existing IMU integration models in \cite{forster2016manifold,henawy2020accurate} which are on the motion of the moving agent rather than the static 3D points of the environment.

The proposed IMU integration model is applied to design a piecewise linear de-skewing algorithm for the 3D points in the $m$th sweep. For simplicity, the transformation matrix from $P_{b,l}(i)$ to $P_{b,l}(j)$ is represented by

{\small \begin{align}
\label{pixelinitial}
T(i,j)=\left[\begin{array}{ll}
F^T(i,j) & t(i,j)\\
0 & 1
\end{array}
\right],
\end{align}}

and the transformation matrix from $P_{b,l}(j)$ to $P_{b,l}(i)$ is

{\small \begin{align}
\label{pixelinitial}
T^{-1}(i,j)=\left[\begin{array}{ll}
F(i,j) & -F(i,j)t(i,j)\\
0 & 1
\end{array}
\right].
\end{align}}

Suppose that the first point and the last point of the $m$th sweep are scanned at the $i_0$th IMU and the $j_0$th IMU. The $l$th point is scanned at $t_{l,m}$. The synchronization point is selected as the $\kappa$th point in the $m$th sweep.  The corresponding IMU is the $\iota$th $(i_0\leq \iota\leq j_0)$ one. Consider the following two cases:

{\it Case 1}: $l$ is smaller than $\iota$. Further assume that there are $q(q<\iota-i_0)$ IMU measurements in the interval $[t_{1,m}, t_{l,m}]$.   It can be easily derived that

{\small \begin{equation}
\label{motioncorrection}
\left[\begin{array}{l}
\hat{X}_l(m) \\ \hat{Y}_l(m) \\ \hat{Z}_l(m)\\1\end{array}
\right]=T(i_0+q, \iota)\left[\begin{array}{l}
X_l(m) \\ Y_l(m) \\ Z_l(m)\\1\end{array}
\right],
\end{equation}}

where the matrix $T(i', \iota)(i_0\leq i'\leq \iota)$ is

{\small \begin{equation}
\label{tranformationmatrix666}
T(i',\iota)=\left[\begin{array}{ll}
F^T(i',\iota) & t(i', \iota)\\
0 & 1\\
\end{array}
\right].
\end{equation}}

{\it Case 2}: $l$ is larger than $\iota$. Further suppose that there are $q'$ IMU measurements in the interval $[t_{\kappa,m}, t_{l,m}]$.
Similarly, it can be derived that

{\small \begin{equation}
\label{motioncorrection1}
\left[\begin{array}{l}
\hat{X}_l(m) \\ \hat{Y}_l(m) \\ \hat{Z}_l(m)\\1\end{array}
\right]=T^{-1}(\iota,\iota+q')\left[\begin{array}{l}
X_l(m) \\ Y_l(m) \\Z_l(m)\\1\end{array}
\right],
\end{equation}}

where the matrix $T^{-1}(\iota,j')$ is

{\small \begin{equation}
\label{tranformationmatrix666666666}
T^{-1}(\iota,j')=\left[\begin{array}{ll}
F(\iota, j') & -F(\iota, j')t(\iota, j')\\
0 & 1\\
\end{array}
\right].
\end{equation}}

Using the above equation, all the points in the $m$th sweep can be aligned with respect to  the synchronization point of the $m$th  sweep. As such, all the motion distortions in the $m$th sweep due to the movement of the agent can be removed. Clearly, the $l$th point is corrected by using all the IMU measurements between the first (or last) point and the point. The proposed de-skewing algorithm does not assume the motion of the agent to be linear during the sweep as required by the de-skewing algorithms in \cite{zhang2014loam,ye2019tightly}.

 Let $b^a(i)$ and $b^{\omega}(i)$ be the biases of $a_b$ and $\omega_b$ in the interval $[i\varsigma, j\varsigma]$. Since the biases of $a_b$ and $\omega_b$ affect the performance of the motion integration model, the values of $a_b(k)$ and $\omega_b(k)$ will be replaced by $(a_b(k)-b^a(i))$ and $(\omega_b(k)-b^{\omega}(i))$, respectively \cite{henawy2020accurate}.   The corrected scans serve as the inputs of an existing LIO such as \cite{ye2019tightly}. The IMU integration model (\ref{motionintegration}) and (\ref{motionintegration2}) can be adopted to derive an IMU integration model of the moving agent which is the same as the model in [16].  The IMU integration model  of the moving agent is then integrated with the LiDAR factor via the joint optimization in \cite{ye2019tightly} with the mean measurement, the covariance matrix and the Jacobian matrix of the proposed model in  \cite{henawy2020accurate}. Due to the space limitation, the details are not repeated in this paper.

\section{Experimental Results}
\label{experimentalresults}

In this section, the proposed de-skewing algorithm is compared with the algorithms in \cite{zhang2014loam}, \cite{ye2019tightly}, \cite{qin2018vins}, \cite{forster2016manifold}, and \cite{shan2020lio}. The proposed algorithm is first implemented on top of the open-source code of LIO\_mapping\footnote{https://github.com/hyye/lio-mapping} in \cite{ye2019tightly}. 

The six datasets with different motion speed and movement in \cite{ye2019tightly} are used to compare all these algorithms. The datasets in \cite{ye2019tightly} were recorded indoor using a Velodyne VLP-16 LiDAR and a Xsens MTi-100 IMU. The sampling rates of the LiDAR and IMU are 5 Hz and 400 Hz, respectively. The toolbox in \cite{zhang2018tutorial} is used to align the results with the ground truth from a motion capture system with a sampling rate of 100 Hz. All the illustrated results are averaged over 20 runs due to the variation in the output estimation. The results are obtained using a desktop with CPU $i9-9900k @ 3.60 GHz \times 16$ and 32 $GiB$ RAM.

\subsection{Comparison of Different Motion Integration Models}
\label{predictedodom}

\begin{table}[!t]
\caption{IMU Odometry For Different Real Sequences With Different Motion Speeds}
\label{RMSE_lio_predicted}
\vspace{-3mm}
\begin{center}
\tabcolsep 0.06in
\renewcommand{\arraystretch}{1.5}
\resizebox{\columnwidth}{!}{%
\begin{tabular}{lcccccccccc}
\hline \hline
\multicolumn{1}{l}{} &
\multicolumn{4}{c}{Proposed Model} & 
\multicolumn{4}{c}{Model in \cite{ye2019tightly}} &
\multicolumn{1}{c}{Difference} & 
\multicolumn{1}{c}{Percentage}  \\
\multicolumn{1}{l}{} &
\multicolumn{2}{c}{Position} & 
\multicolumn{2}{c}{Orientation} &
\multicolumn{2}{c}{Position} & 
\multicolumn{2}{c}{Orientation} &
\multicolumn{1}{c}{Position} & 
\multicolumn{1}{c}{}  \\
  &  RMSE & $\sigma$ & RMSE & $\sigma$   & RMSE & $\sigma$  &  RMSE & $\sigma$  &  RMSE &   \\ \hline \hline
\rowcolor{Lightcyan}
Sequence &   [cm]& [cm] & [deg]& [deg] &  [cm]& [cm] & [deg]& [deg]  & [cm] & [$\%$]  \\

Fast 1 &  \cellcolor{Lightyellow} \textbf{11.96} & 4.37 & 0.1260 & 0.0971 & 16.25 & 4.79 & \cellcolor{Lightyellow} \textbf{0.1245} & 0.1043  & 4.29  & 26.40  \\

Fast 2 &  \cellcolor{Lightyellow} \textbf{34.60} & 17.42 & 0.2361 & 0.0775 & 37.68 & 16.64 & \cellcolor{Lightyellow} \textbf{0.1401} & 0.0645  & 3.08  & 8.17  \\

Med 1 &  \cellcolor{Lightyellow} \textbf{26.17} & 14.07 & 0.0852 & 0.0301 & 27.05 & 14.75 & \cellcolor{Lightyellow} \textbf{0.0785} & 0.0318  & 0.88  & 3.25  \\

Med 2 &  \cellcolor{Lightyellow} \textbf{31.08} & 13.57 & \cellcolor{Lightyellow} \textbf{0.3481} & 0.3080 & 32.81 & 14.72 & 0.3698 & 0.3080  & 1.73  & 5.27  \\

Slow 1 &  45.56 & 23.93 & 0.3778 & 0.1560 & \cellcolor{Lightyellow} \textbf{45.25} & 23.47 & \cellcolor{Lightyellow} \textbf{0.3749} & 0.1529  & -0.31  & -0.68   \\

Slow 2 &  46.53 & 22.92 & 0.2984 & 0.1975 & \cellcolor{Lightyellow} \textbf{44.75} & 19.15 & \cellcolor{Lightyellow} \textbf{0.2950} & 0.2069  & -1.78  & -3.98  \\ \hline \hline
\end{tabular}%
}
\end{center}
\end{table}

\begin{table*}[!t]
\caption{Comparison of Proposed LIO Algorithm with the LIO Algorithm in \cite{ye2019tightly} without Metric Maps \\(RMSE On Position, Orientation and Their Standard Deviation ($\sigma$)  For Different Real Sequences With Different Motion Speeds)}
\vspace{-3mm}
\begin{center}
\tabcolsep 0.06in
\renewcommand{\arraystretch}{1.2}
\begin{tabular}{lcccccccccccc}
\hline \hline
\multicolumn{1}{l}{} &
\multicolumn{4}{c}{Proposed LIO} & 
\multicolumn{4}{c}{LIO in \cite{ye2019tightly}} &
\multicolumn{1}{c}{Difference} & 
\multicolumn{1}{c}{Percentage} & 
\multicolumn{1}{c}{Total Distance} & 
\multicolumn{1}{c}{Linear Velocity} \\
\multicolumn{1}{l}{} &
\multicolumn{2}{c}{Position} & 
\multicolumn{2}{c}{Orientation} &
\multicolumn{2}{c}{Position} & 
\multicolumn{2}{c}{Orientation} &
\multicolumn{1}{c}{Position} & 
\multicolumn{1}{c}{} & 
\multicolumn{1}{c}{} & 
\multicolumn{1}{c}{} \\
  &  RMSE & $\sigma$ & RMSE & $\sigma$   & RMSE & $\sigma$  &  RMSE & $\sigma$  &  RMSE &   &  &  \\ \hline \hline
\rowcolor{Lightcyan}
Sequence &   [cm]& [cm] & [deg]& [deg] &  [cm]& [cm] & [deg]& [deg]  & [cm] & [$\%$]  & [m] & [m/s]\\

Fast 1 &  \cellcolor{Lightyellow} \textbf{9.13} & 4.46 & 0.1781 & 0.1602 & 11.79 & 5.50 & \cellcolor{Lightyellow} \textbf{0.1507} & 0.1327  & 2.66  & 22.56   & 31.59 & 0.8904\\

Fast 2 &  \cellcolor{Lightyellow} \textbf{11.94} & 6.09 & 0.2322 & 0.2098 & 14.47 & 7.63 & \cellcolor{Lightyellow} \textbf{0.2034} & 0.1747  & 2.53  & 17.48   & 33.12  & 0.9404\\

Med 1 &  \cellcolor{Lightyellow} \textbf{24.91} & 13.18 & 0.0787 & 0.0582 & 25.87 & 13.21 & \cellcolor{Lightyellow} \textbf{0.0781} & 0.0568  & 0.96  & 3.71   & 38.76 & 0.7047\\

Med 2 &  \cellcolor{Lightyellow} \textbf{22.06} & 11.25 & \cellcolor{Lightyellow} \textbf{0.3496} & 0.3302 & 24.39 & 11.87 & 0.3583 & 0.3358  & 2.33  & 9.55   & 40.33 & 0.8528\\

Slow 1 &  \cellcolor{Lightyellow} \textbf{17.69} & 9.64 & 0.2870 & 0.2744 & 18.01 & 9.91 & \cellcolor{Lightyellow} \textbf{0.2839} & 0.2707  & 0.32  & 1.78   & 23.02  &  0.6511\\

Slow 2 &  \cellcolor{Lightyellow} \textbf{12.21} & 6.70 & \cellcolor{Lightyellow} \textbf{0.3251} & 0.3139 & 14.17 & 7.72 & 0.3252 & 0.3131  & 1.96  & 13.83   & 24.43 & 0.6950\\ \hline \hline

\end{tabular}%
\end{center}
\label{RMSETable_lio}
\end{table*}

This subsection aims to compare the closed form motion integration of the moving agent with the IMU integration model in \cite{forster2016manifold,qin2018vins} because the LIO\_mapping uses the discrete IMU model in \cite{forster2016manifold,qin2018vins}. Note that the predicted IMU integration in the \cite{ye2019tightly} was used in this evaluation. The IMU integration period is $0.2$ $sec$.

The results in Table \ref{RMSE_lio_predicted} are obtained using a sampling rate of 100 Hz  for both the IMU odometry and ground truth as the default in the open-source code. The IMU odometry  has been aligned with the ground truth using the first 50 estimated poses under the evaluation trajectory toolbox in \cite{zhang2018tutorial}. The results show that the proposed model outperforms the state-of-the-art one for complicated motion conditions. For example, the proposed model outperforms the state-of-the-art one on Fast 1 and Fast 2 sequences by $26\%$ and $8.2\%$, respectively. The reason is that the proposed model takes into consideration the high dynamic change of the motion. On the contrary, the state-of-the-art one slightly outperforms the proposed model on Slow 1 and Slow 2 sequences. That is because the high sampling rate of IMU mitigates its discretization effect. The results are consistent with the evaluation in \cite{henawy2019accurate,henawy2020accurate}.

\subsection{Evaluation of De-skewing and LIO without Metric Maps}

In this subsection, the proposed de-skewing algorithm and the closed form motion integration model are evaluated using the LIO framework  without enabling any metric map. The proposed de-skewing algorithm (\ref{motioncorrection})-(\ref{tranformationmatrix666}) is also implemented  in the open-source code \cite{ye2019tightly}. The proposed motion integration model is integrated with  the LiDAR factor via a joint optimization  with the mean measurement, the covariance matrix and the Jacobian matrix of the proposed model in  \cite{henawy2020accurate}. Note that all the estimated states are aligned with the ground truth using the first 50 estimation poses under the toolbox in \cite{zhang2018tutorial}. The sampling rate of the LiDAR is 5 Hz and all other settings in the experiment are kept the same for the fair comparison with the LIO algorithm in \cite{ye2019tightly}.

As shown in Table \ref{RMSETable_lio}, the proposed LIO algorithm outperforms the state-of-the-art \cite{ye2019tightly} by $22.49\%$ and $13.79\%$ on Fast 1 and Slow 2 sequences, respectively. The is because the proposed motion integration model has good representation for the IMU excitation and takes into consideration the high dynamic change between two consecutive IMU measurements for the fast motion. This conclusion is foreseeable and consistent with the results of the IMU odometry test in Table \ref{RMSE_lio_predicted}.

\begin{figure}[!t]
	\centering{
\subfigure[Fast 1]{\includegraphics[width=2.9in]{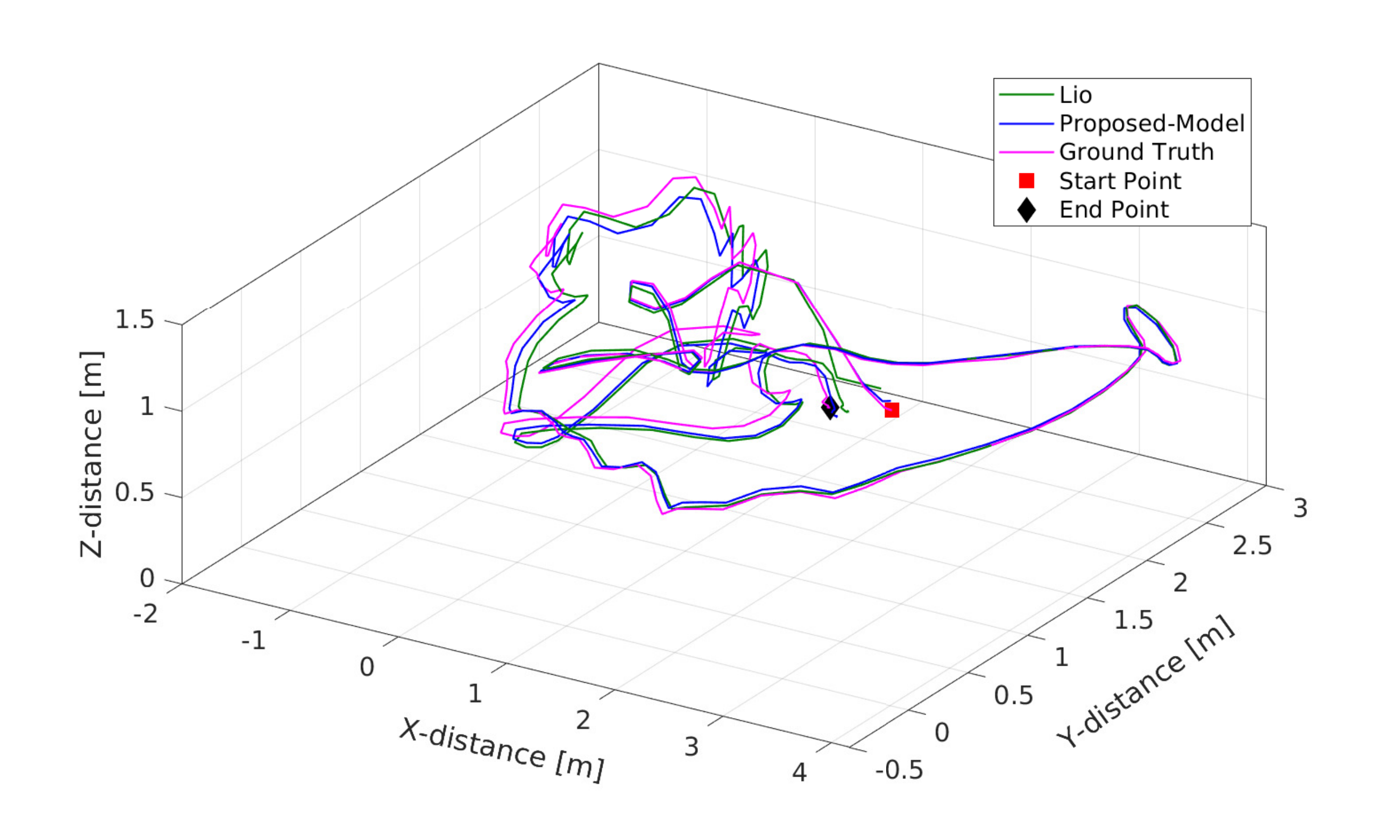}}}
	\centering{
\subfigure[Slow 2]{\includegraphics[width=2.9in]{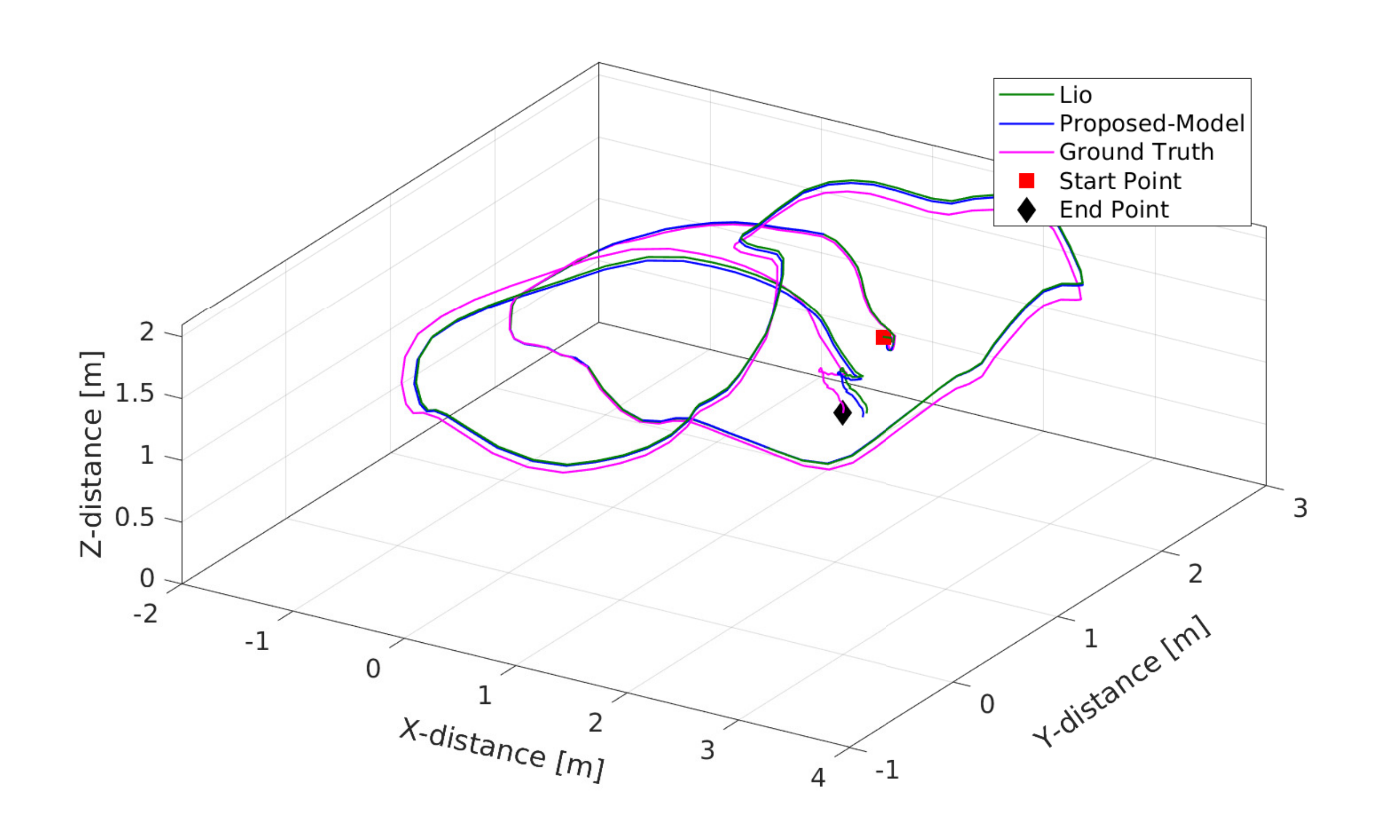}}}
\caption{Estimated trajectories against the ground-truth trajectories}
\label{Trajectory_fast1_slow2}
    \vspace{-0.2cm}
\end{figure}

\begin{figure}[!t]
	\centering{
\subfigure[RMSE on Position Error]{\includegraphics[width=3.3in]{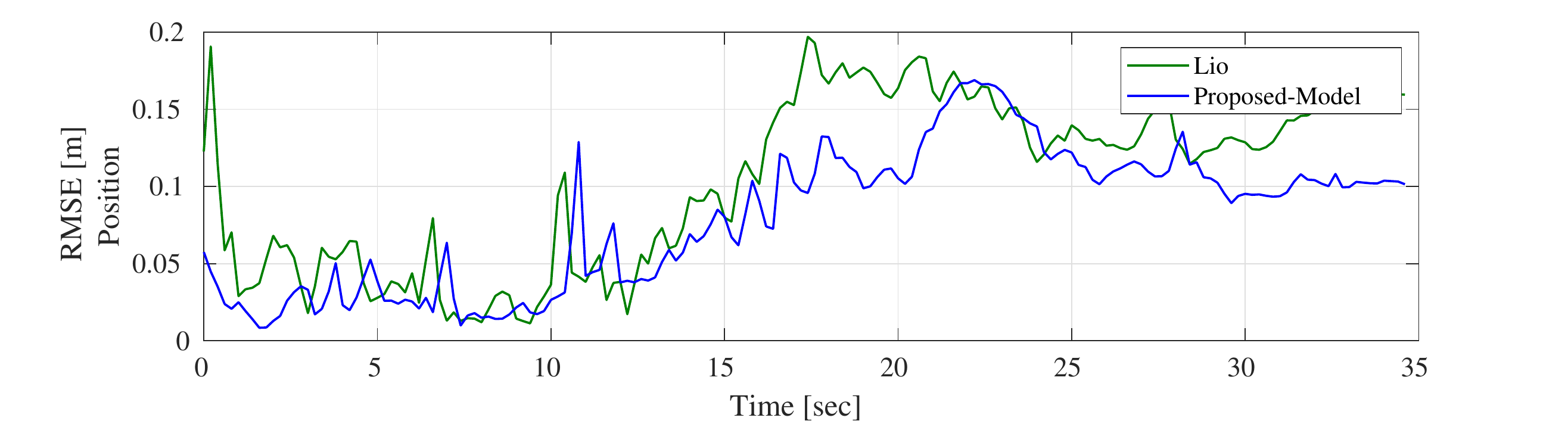}}}
	\centering{
\subfigure[RMSE on Orientation Error]{\includegraphics[width=3.3in]{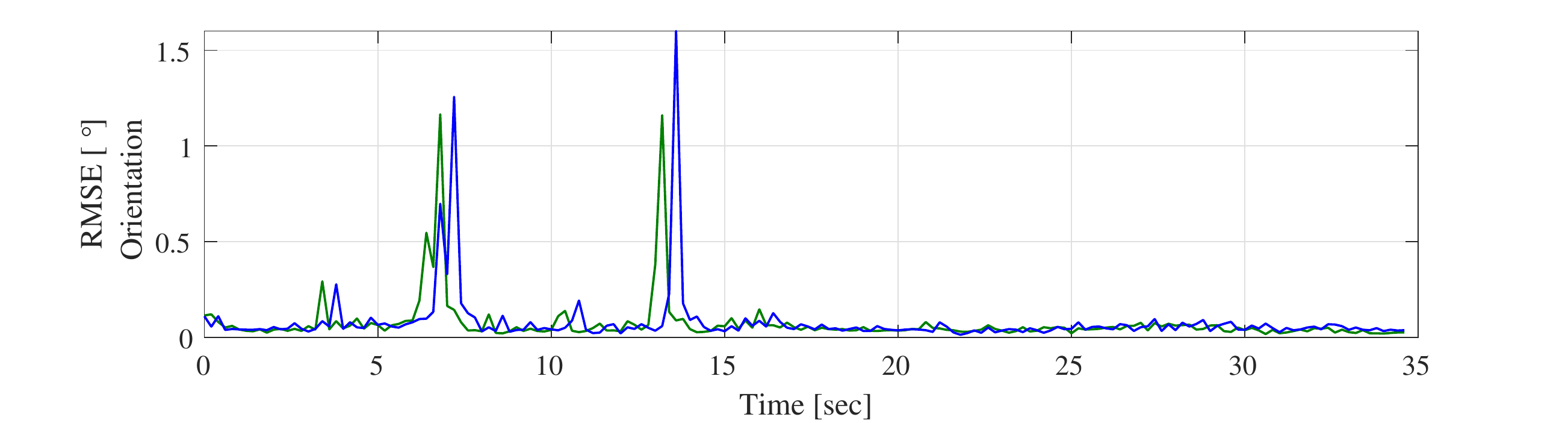}}}
	\centering{
\subfigure[Absolute Error]{\includegraphics[width=3.3in]{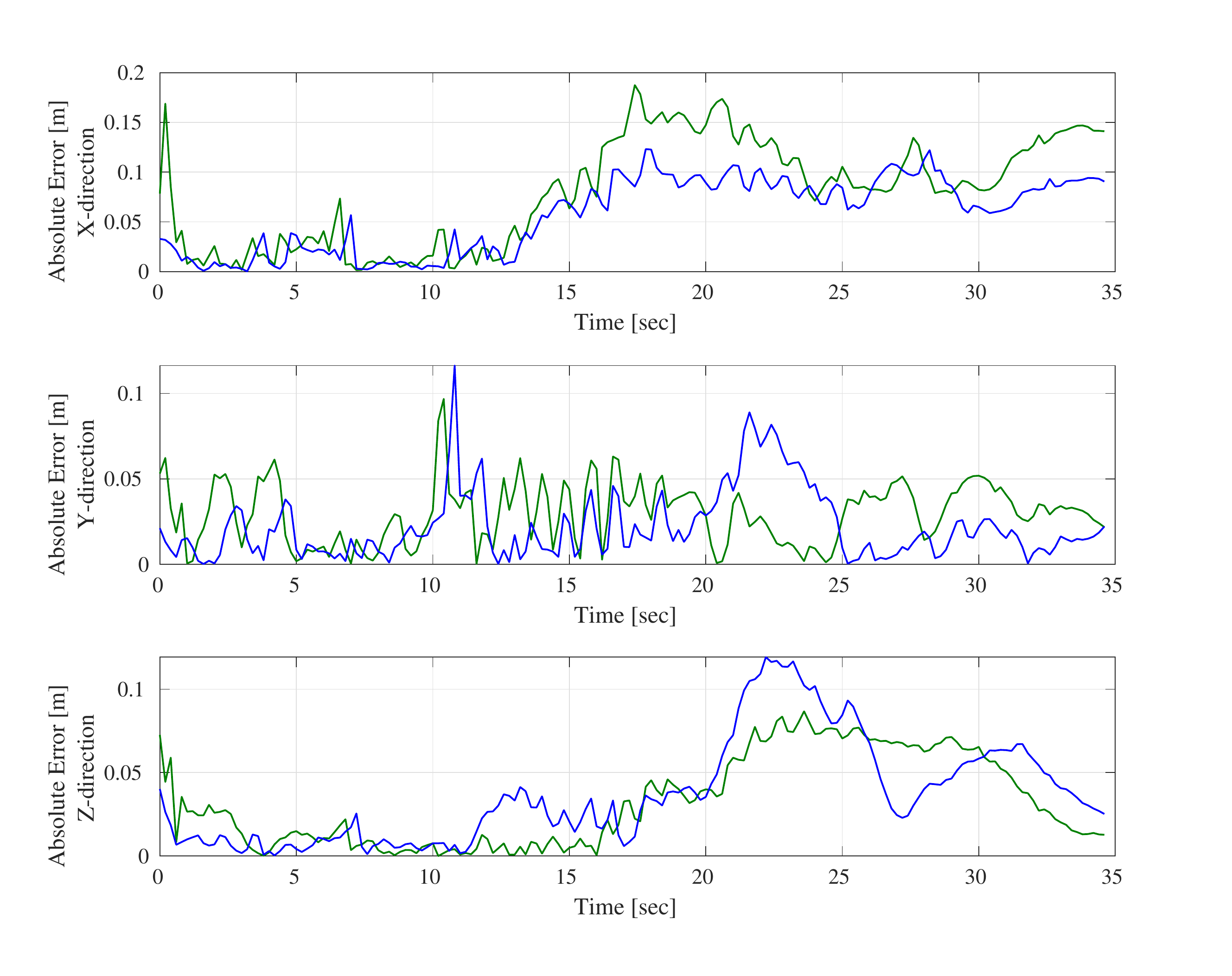}}}
\caption[RMSE and position analysis on Fast 1 Sequence for the proposed model and LIO model]{RMSE and position analysis on Fast 1 Sequence for the proposed LIO and the LIO \cite{ye2019tightly}}
\label{analysis_fast1}
    \vspace{-0.2cm}
\end{figure}

\begin{figure}[!t]
	\centering{
\subfigure[RMSE on Position Error]{\includegraphics[width=3.3in]{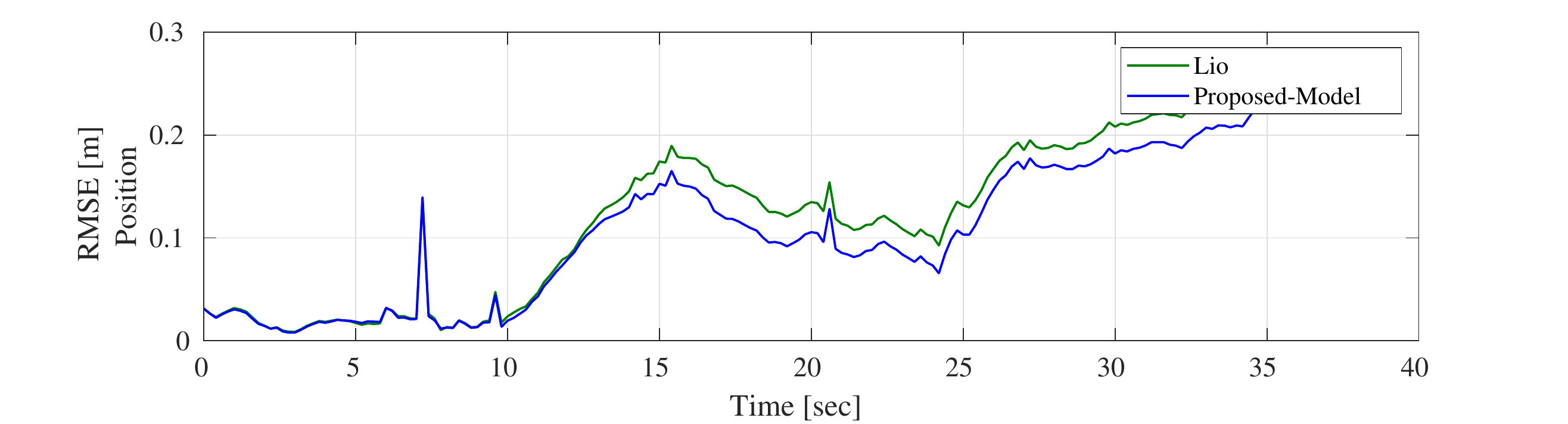}}}
	\centering{
\subfigure[RMSE on Orientation Error]{\includegraphics[width=3.3in]{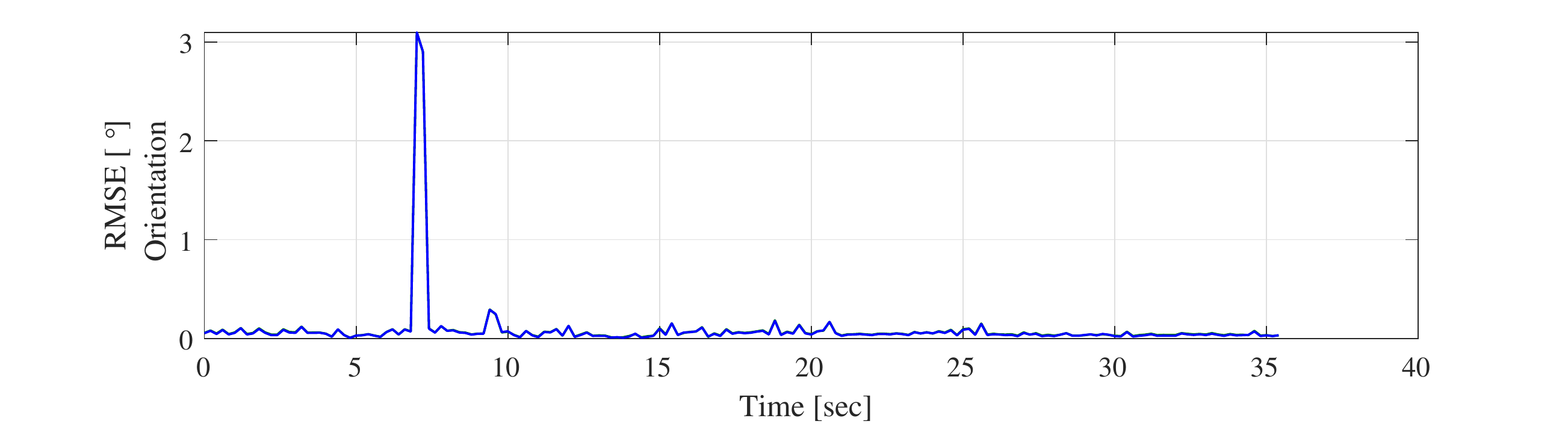}}}
	\centering{
\subfigure[Absolute Error]{\includegraphics[width=3.3in]{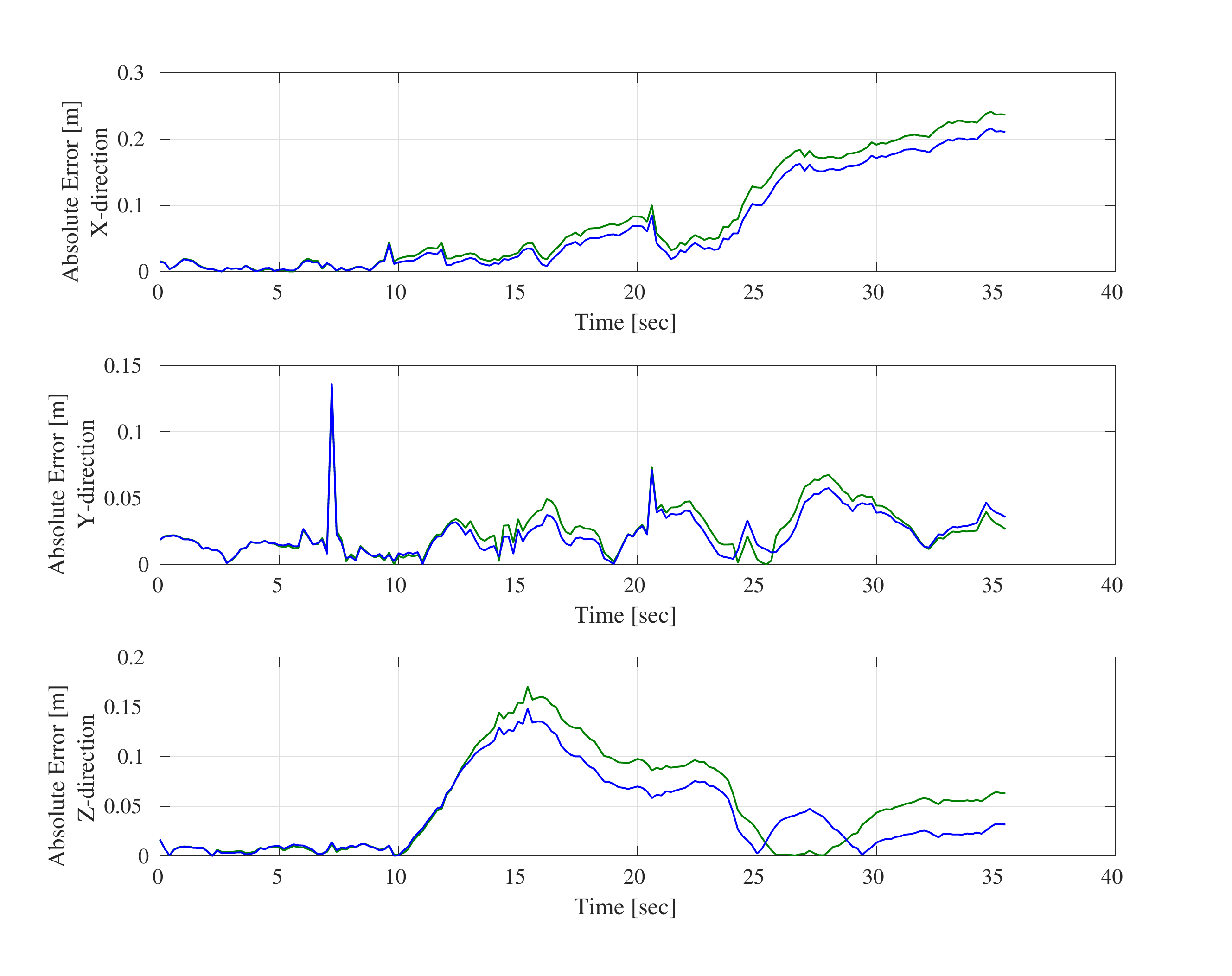}}}
\caption[RMSE and position analysis on Slow 2 Sequence for the proposed model and LIO]{RMSE and position analysis on Slow 2 Sequence for the proposed LIO and the LIO \cite{ye2019tightly}}
\label{analysis_slow2}
    \vspace{-0.2cm}
\end{figure}

In the case of the slow motion, the proposed  algorithm outperforms the state-of-the-art one in \cite{ye2019tightly} by $1.8\%$ and $13.79\%$ on Slow 1 and Slow 2 sequences, respectively. The results seem to be unexpected because the state-of-the-art motion integration model in  \cite{forster2016manifold,qin2018vins} outperforms the proposed motion integration model in the IMU odometry as shown in Table \ref{RMSE_lio_predicted}. The  improvement comes from the proposed IMU model which has been deployed as a tightly coupled with LiDAR factor, and from the transformation matrix (\ref{tranformationmatrix666}) or  (\ref{tranformationmatrix666666666}) which is more accurate than the corresponding matrix in the state-of-the-art one. The de-skewing algorithm in \cite{ye2019tightly} first computes the transformation matrix from the first point to the last point, and then linearly interpolates a transformation matrix for each point to be corrected. The interpolated transformation matrix is not accurate if the motion of the agent is not linear.

Figure \ref{Trajectory_fast1_slow2} shows the aligned estimated trajectories against the ground-truth for Fast 1 and Slow 2 sequences. Clearly, the Fast 1 trajectory is more complicated and has many turning points than the Slow 2 trajectory. The proposed algorithm behaves well on both the trajectories. Figures \ref{analysis_fast1} and \ref{analysis_slow2} show the RMSE on position and orientation, and the absolute error analysis in the three dimensions on Fast 1 and Slow 2 Datasets, respectively. It shows that the proposed algorithm has a good performance in terms of the RMSE over the full trajectory. It is also observed that the difference in state estimation between the proposed algorithm and the state-of-the-art one in \cite{ye2019tightly} is increased along the travelling time. Moreover, Table \ref{RMSETable_lio} shows the total travelling distance and linear velocity for each sequence.

Overall, the proposed LIO algorithm outperforms the state-of-the-art in \cite{ye2019tightly} especially for complicated movements. The proposed model shows the reliability and robustness on the state estimation.

\subsection{Evaluation of De-skewing and LIO with Metric Maps}

In this subsection,  effect of metric maps to the LIO will be studied and discussed. The proposed algorithm is compared with the state-of-the-art one in \cite{ye2019tightly} and the LOAM \footnote{https://github.com/laboshinl/loam\_velodyne} \cite{zhang2014loam}. LOAM is one of the top accurate state estimation algorithms, and it shows a high accuracy estimation on the most popular dataset in the LO field, i.e., Kitti datasets\footnote{http://www.cvlibs.net/datasets/kitti/eval\_odometry.php} \cite{geiger2013vision}. This dataset has been recorded and obtained using a ground vehicle (GV). It is worth noting that the state estimation using the LOAM is done without any assistance from the IMU.

\begin{table}[t]
\caption{Comparison of Proposed LIO Algorithm with the LIO Algorithm in \cite{ye2019tightly} with Metric Maps and the LOAM in \cite{zhang2014loam}(RMSE On Position For Different Real Sequences With Different Motion Speeds)}
\vspace{-3mm}
\begin{center}
\tabcolsep 0.06in
\renewcommand{\arraystretch}{1.5}
\resizebox{\columnwidth}{!}{%
\begin{tabular}{lcccccccccc}
\hline \hline
\multicolumn{1}{l}{} &
\multicolumn{2}{c|}{Proposed LIO with Metric Map} &
\multicolumn{4}{c|}{LIO in \cite{ye2019tightly} with Metric Map} &
\multicolumn{4}{c}{LOAM \cite{zhang2014loam}} \\
\multicolumn{1}{l}{} &
\multicolumn{2}{c|}{Position} &
\multicolumn{4}{c|}{Position} &
\multicolumn{4}{c}{Position} \\
\multicolumn{1}{l}{} &
\multicolumn{1}{c}{RMSE} &
\multicolumn{1}{c|}{$\sigma$} &
\multicolumn{1}{c}{RMSE} &
\multicolumn{1}{c}{$\sigma$} &
\multicolumn{1}{c}{Difference} &
\multicolumn{1}{c|}{Percentage} &
\multicolumn{1}{c}{RMSE} &
\multicolumn{1}{c}{$\sigma$} &
\multicolumn{1}{c}{Difference} &
\multicolumn{1}{c}{Percentage} \\
\hline \hline
\rowcolor{Lightcyan}
Sequence  & [cm]  & [cm] &  [cm] & [cm] & [cm] & [$\%$] & [cm] &  [cm] & [cm]  & [$\%$] \\
Fast 1 &  \cellcolor{Lightyellow}\textbf{5.03} & 2.10  &  5.09 & 2.03  & 0.06 &  1.05 & \cellcolor{Green} 155.49 & \cellcolor{Green} 69.37 &  \cellcolor{Red} -  &  \cellcolor{Red} -   \\
Fast 2 &  \cellcolor{Lightyellow} \textbf{9.65}& 5.75 & 9.92 & 5.75  & 0.27 & 2.71
 & \cellcolor{Green} 152.81  & \cellcolor{Green} 102.15  & \cellcolor{Red} -   & \cellcolor{Red} -  \\

Med 1 &  10.94 & 5.32   &  10.94  & 5.24 & - & -
 & \cellcolor{Green} 203.78  & \cellcolor{Green} 136.82  & \cellcolor{Red} -  &  \cellcolor{Red} - \\

Med 2 &  \cellcolor{Lightyellow} \textbf{7.74}& 6.19 & 7.97 &  6.33& 0.23 & 3.00
 &\cellcolor{Green} 158.93 & \cellcolor{Green} 115.31   & \cellcolor{Red} -   & \cellcolor{Red} -  \\

Slow 1 &  6.09 & 2.77 & \cellcolor{Lightyellow} \textbf{5.80} &   2.67 & -0.29 & -4.97  &  15.57  & 8.24 &  9.48   & 60.89   \\

Slow 2 &  \cellcolor{Lightyellow} \textbf{5.13} & 2.20& 5.32 & 2.48 & 0.19 & 3.57 &  9.22  & 5.22  & 4.09  & 44.36  \\ \hline \hline

\end{tabular}%
}
\end{center}
\label{RMSETable_map}
\end{table}

\begin{figure}[!b]
\centering{
\includegraphics[width=2.9in]{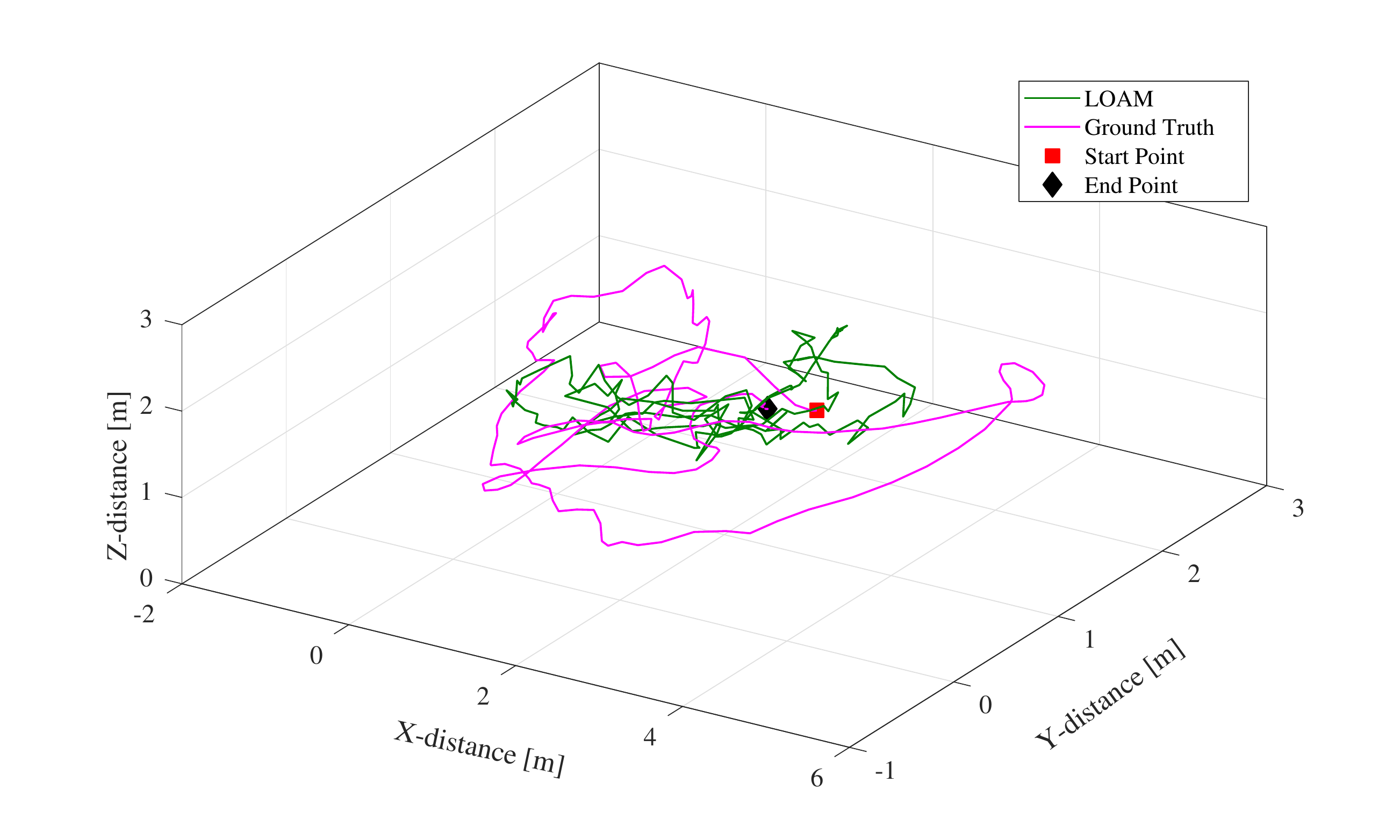}}
\caption{Estimated trajectory via the LOAM against the ground-truth one on Fast 1 Dataset}
\label{Trajectory_fast1_loam}
\end{figure}

Table \ref{RMSETable_map} illustrates the RMSE on position of the full trajectory for the proposed algorithm,  the  LIO\_mapping in \cite{ye2019tightly}, and the LOAM. The results show that the LOAM fails to estimate the state in the presence of fast or medium speed motion. There is no assistance from the IMU  and the LOAM fails on these conditions. The RMSE of position for the LOAM on the  Fast and Med datasets are highlighted using the green color in Table \ref{RMSETable_map}. The red highlight indicates that the LOAM fails to estimation the position of the robot as shown in Figure \ref{Trajectory_fast1_loam}.

The proposed algorithm shows a superior's performance compared to the LOAM on these conditions by $60.89\%$ on Slow 1 dataset and by $44.36\%$ on Slow 2 dataset. This is because 1) the proposed de-skewing algorithm does not assume the motion of
the agent to be linear during the sweep as required by the LOAM; and 2) the proposed motion integration model is integrated with  the LiDAR factor via the tightly coupled joint optimization in \cite{ye2019tightly}. It should be pointed out that the accuracy of LOAM will be improved if an IMU is deployed as an additional sensor as mentioned in \cite{zhang2017low}.

The results in Table \ref{RMSETable_map} show that the proposed algorithm and the state-of-the-art one in \cite{ye2019tightly} are comparable in presence of metric maps. This is because the results  highly depend on the metric map with salient features.

\subsection{Further Evaluation of The Proposed De-skewing Algorithm}
\label{lio_sam}

Recently, an interesting LIO algorithm was proposed in \cite{shan2020lio}. The de-skewing algorithm in \cite{shan2020lio} is replaced by the proposed one and the implementation is on top of the open-source code of LIO-SAM\footnote{https://github.com/TixiaoShan/LIO-SAM} and one dataset in \cite{shan2020lio}. Both GPS and loop closure are disabled to test the effectiveness of the proposed model.

One dataset in \cite{shan2020lio} was recorded in the park which is covered by vegetation using a Velodyne VLP-16 LiDAR and a MicroStrain 3DM-GX5-25 IMU mounted on UGV. The sampling rates of the LiDAR and IMU are 10 Hz and 500 Hz, respectively.

\begin{figure}[htb]
	\centering{
\includegraphics[width=2.9in]{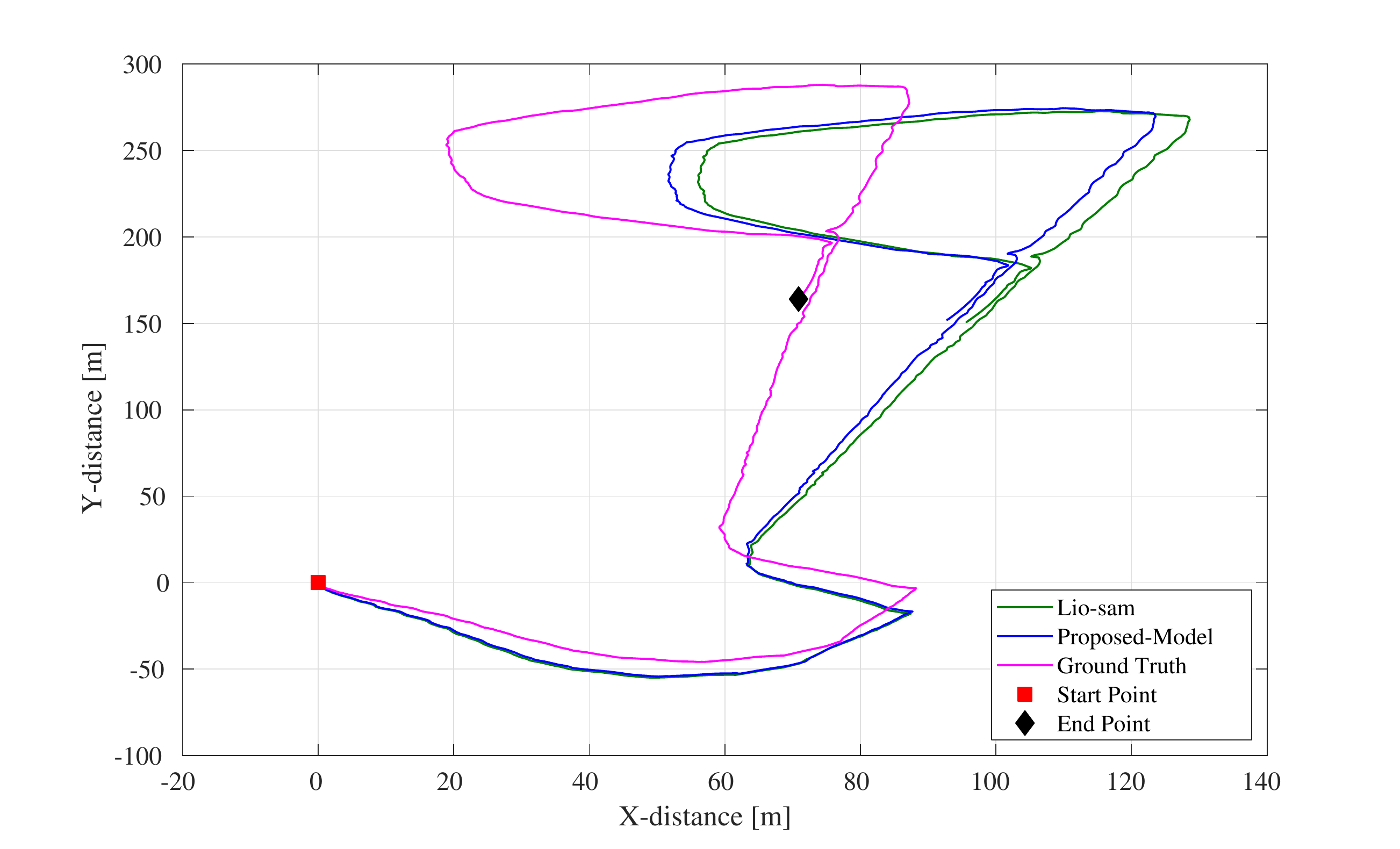}}
\caption{Estimated trajectories against the ground-truth trajectories}
\label{Trajectory_park}
    \vspace{-0.2cm}
\end{figure}

\begin{figure}[htb]
	\centering{
\subfigure[RMSE on Position Error]{\includegraphics[width=3.3in]{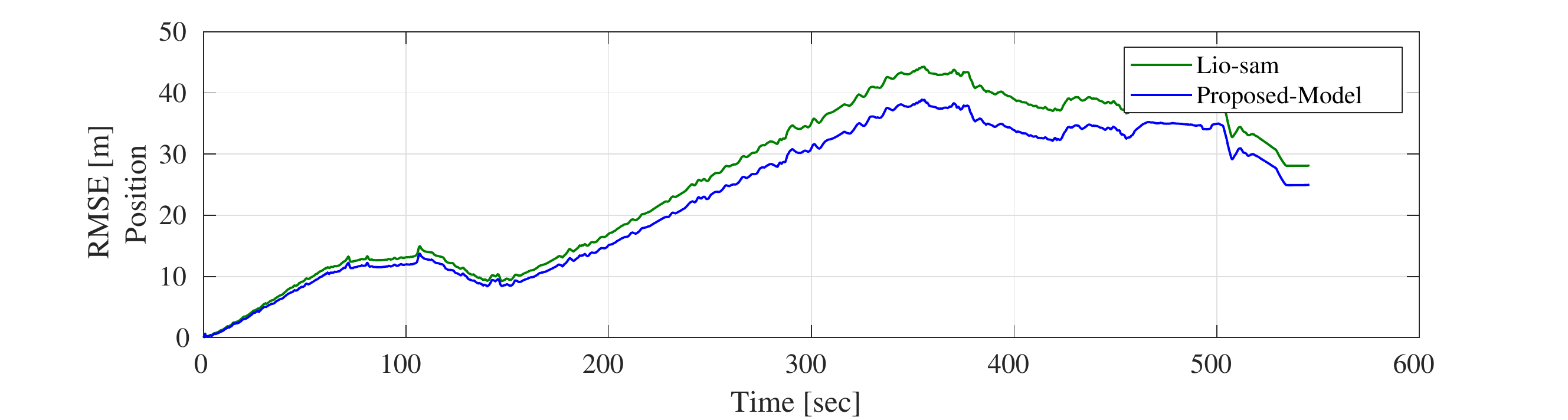}}}
	\centering{
\subfigure[Absolute Error]{\includegraphics[width=3.3in]{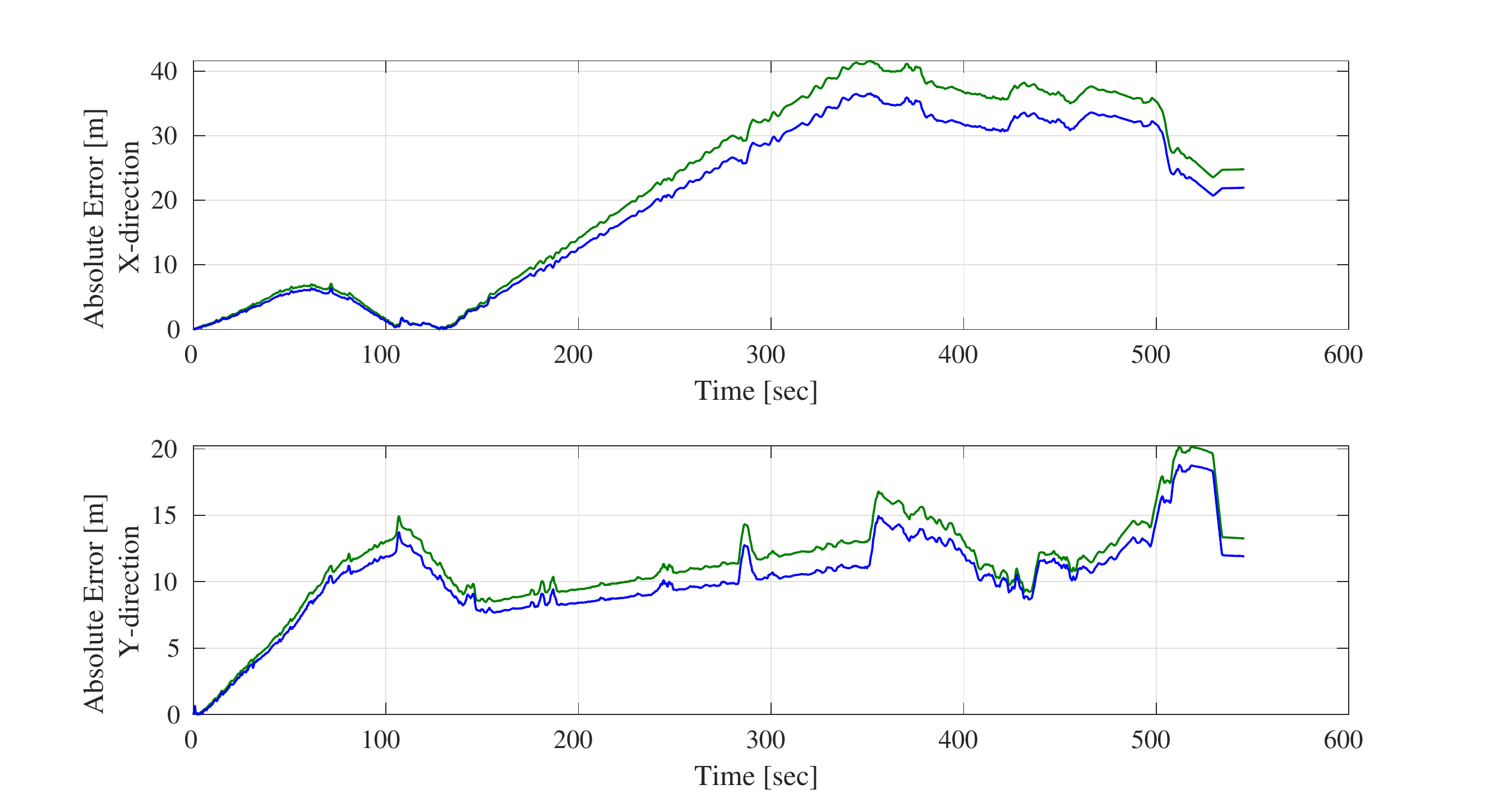}}}
\caption{RMSE and position analysis on Park dataset}
\label{analysis_park}
    \vspace{-0.2cm}
\end{figure}

Table \ref{RMSE_2D_gps} shows the Root Mean Square Error (RMSE) w.r.t the ground-truth provided by GPS. The RMSE results obtained in Table \ref{RMSE_2D_gps} do not include the error in the z-direction or orientation errors. The reason is that the ground-truth in the park dataset does not have z-direction or orientation measurements. The results show that the proposed algorithm outperforms the state-of-the-art LIO-SAM by $11.53\%$. Figure \ref{Trajectory_park} shows the estimated trajectories against the ground-truth on the park dataset, while Figures \ref{analysis_park}.(a) and \ref{analysis_park}.(b) show the RMSE on 2D position and the absolute error analysis in the two dimensions on the park dataset, respectively. It shows that the proposed algorithm has a slightly better performance and estimation regarding the RMSE over the full trajectory w.r.t the state-of-the-art one.

\begin{table}[!t]
\caption{RMSE On 2D Position w.r.t GPS}
\label{RMSE_2D_gps}
\vspace{-3mm}
\begin{center}
\tabcolsep 0.06in
\renewcommand{\arraystretch}{1.5}
\resizebox{\columnwidth}{!}{%
\begin{tabular}{lcccccccc}
\hline \hline
\multicolumn{1}{l}{} &
\multicolumn{2}{c}{Proposed} & 
\multicolumn{2}{c}{Lio\_sam \cite{shan2020lio}} &
\multicolumn{1}{c}{Difference} & 
\multicolumn{1}{c}{Percentage} & 
\multicolumn{1}{c}{Total Distance} & 
\multicolumn{1}{c}{Linear Velocity} \\
\multicolumn{1}{l}{} &
\multicolumn{1}{c}{RMSE} & 
\multicolumn{1}{c}{$\sigma$} &
\multicolumn{1}{c}{RMSE} & 
\multicolumn{1}{c}{$\sigma$} &
\multicolumn{4}{c}{}  \\
  \hline \hline
\rowcolor{Lightcyan}
Sequence & [m] & [m] &  [m]& [m]& [m]&  [$\%$] & [m] & [m/s] \\

park &  \cellcolor{Lightyellow} \textbf{25.63} &  11.56 & 28.97 &  13.21  & 3.34 & 11.53 & 656.85  & 1.1932\\ \hline \hline
\end{tabular}%
}
\end{center}
\end{table}

\section{Conclusion Remarks}
\label{conclusion}

A novel piecewise linear de-skewing algorithm has been proposed for  LiDAR inertial odometry (LIO) of fast moving agents using high frequency motion information provided by an inertial measurement unit (IMU). The robustness of the LIO can be enhanced by incorporating the proposed de-skewing algorithm into the LIO. Experimental results validated the performance of the proposed de-skewing algorithm. Due to its simplicity, the proposed de-skewing algorithm is practical for real world deployment.

\section*{Acknowledgement}
This research is supported by SERC grant No. 192 25 00049 from the National Robotics Programme (NRP).


\bibliography{IEEEabrv,references}
\bibliographystyle{IEEEtran}

\end{document}